\documentclass[10pt,twocolumn,letterpaper]{article}

\pdfoutput=1
\usepackage{cvpr}
\usepackage{times}
\usepackage{epsfig}
\usepackage{graphicx}
\usepackage{amsmath}
\usepackage{amssymb}

\usepackage[breaklinks=true,bookmarks=false]{hyperref}

\cvprfinalcopy 

\def\cvprPaperID{****} 

\begin{document}

\title{ Surpassing Human-Level Face Verification Performance on LFW with GaussianFace}

\author{Chaochao Lu~~~~~~~~~~~~~~~~~~~~~Xiaoou Tang\\
{\small Department of Information Engineering, The Chinese University of Hong Kong}\\
{\tt\small \{lc013, xtang\}@ie.cuhk.edu.hk}
}

\maketitle

\begin{abstract}
Face verification remains a challenging problem in very complex conditions with large variations such as pose, illumination, expression, and occlusions. This problem is exacerbated when we rely unrealistically on a single training data source, which is often insufficient to cover the intrinsically complex face variations. This paper proposes a principled multi-task learning approach based on Discriminative Gaussian Process
Latent Variable Model, named GaussianFace, to enrich the diversity of training data. In comparison to existing methods, our model exploits additional data from multiple source-domains to improve the generalization performance of face verification in an unknown target-domain. Importantly, our model can adapt automatically to complex data distributions, and therefore can well capture complex face variations inherent in multiple sources. Extensive experiments demonstrate the effectiveness of the proposed model in learning from diverse data sources and generalize to unseen domain. Specifically, the accuracy of our algorithm achieves an impressive accuracy rate of 98.52\% on the well-known and challenging Labeled Faces in the Wild (LFW) benchmark \cite{LFWTech}. For the first time, the human-level performance in face verification (97.53\%) \cite{kumar2009attribute} on LFW is surpassed. \footnote{For project update, please refer to mmlab.ie.cuhk.edu.hk.}
\end{abstract}

\section{Introduction}

Face verification, which is the task of determining whether a pair of face images are from the same person, has been an active research topic in computer vision for decades \cite{kumar2009attribute,Huang2012Learning,seo2011face,berg2012tom,simonyan2004fisher,liprobabilistic,Dong2013Blessing,Cao2013Transfer}. It has many important applications, including surveillance, access control, image retrieval, and automatic log-on for personal computer or mobile devices. However, various visual complications deteriorate the performance of face verification, as shown by numerous studies on real-world face images from the wild \cite{LFWTech}. The Labeled Faces in the Wild (LFW) dataset is well known as a challenging benchmark for face verification. The dataset provides a large set of relatively unconstrained face images with complex variations in pose, lighting, expression, race, ethnicity, age, gender, clothing, hairstyles, and other parameters. Not surprisingly, LFW has proven difficult for automatic face verification methods \cite{LFWTech,kumar2009attribute}. Although there has been significant work \cite{Huang2012Learning,Cao2013Transfer,berg2012tom,Dong2013Blessing,simonyan2004fisher,chen2012bayesian,yin2011associate,Sun2013Hybrid,sun2014,deepFace} on LFW and the accuracy rate has been improved from 60.02\% \cite{turk1991face} to 97.35\% \cite{deepFace} since LFW is established in 2007, these studies have not closed the gap to human-level performance \cite{kumar2009attribute} in face verification.

Why could not we surpass the human-level performance? Two possible reasons are found as follows:

1) Most existing face verification methods assume that the training data and the test data are drawn from the same feature space and follow the same distribution. When the distribution changes, these methods may suffer a large performance drop \cite{wright2009implicit}. However, many practical scenarios involve cross-domain data drawn from different facial appearance distributions. Learning a model solely on a single source data often leads to overfitting due to dataset bias \cite{torralba2011unbiased}. Moreover, it is difficult to collect sufficient and necessary training data to rebuild the model in new scenarios, for highly accurate face verification specific to the target domain. In such cases, it becomes critical to exploit more data from multiple source-domains to improve the generalization of face verification methods in the target-domain.

2) Modern face verification methods are mainly divided into two categories: extracting low-level features \cite{lowe2004distinctive,ahonen2006face,liu2002gabor,cao2010face, hussain2012face}, and building classification models \cite{Zhu2013Deep,Sun2013Hybrid,chen2012bayesian,moghaddam2000bayesian,liprobabilistic,turk1991face,berg2012tom,kumar2009attribute,simonyan2004fisher,li2005nonparametric}.
Although these existing methods have made great progress in face verification, most of them are less flexible when dealing with complex data distributions. For the methods in the first category, for example, low-level features such as SIFT \cite{lowe2004distinctive}, LBP \cite{ahonen2006face}, and Gabor \cite{liu2002gabor} are handcrafted. Even for features learned from data \cite{cao2010face, hussain2012face}, the algorithm parameters (such as the depth of random projection tree, or the number of centers in k-means) also need to be specified by users. Similarly, for the methods in the second category, the architectures of deep networks in \cite{Zhu2013Deep,Sun2013Hybrid,ping2014,sun2014} (for example, the number of layers, the number of nodes in each layer, etc.), and the parameters of the models in \cite{liprobabilistic,berg2012tom,kumar2009attribute,simonyan2004fisher} (for example, the number of Gaussians, the number of classifiers, etc.) must also be determined in advance. Since most existing methods require some assumptions to be made about the structures of the data, they cannot work well when the assumptions are not valid. Moreover, due to the existence of the assumptions, it is hard to capture the intrinsic structures of data using these methods.

To this end, we propose the Multi-Task Learning approach based on Discriminative Gaussian Process Latent Variable Model (DGPLVM) \cite{urtasun2007discriminative}, named \emph{GaussianFace}, for face verification. Unlike most existing studies \cite{Huang2012Learning,berg2012tom,Dong2013Blessing,simonyan2004fisher,chen2012bayesian} that rely on a single training data source, in order to take advantage of more data from multiple source-domains to improve the performance in the target-domain, we introduce the multi-task learning constraint to DGPLVM. Here, we investigate the asymmetric multi-task learning because we only focus on the performance improvement of the target task. From the perspective of information theory, this constraint aims to maximize the mutual information between the distributions of target-domain data and multiple source-domains data. Moreover, the GaussianFace model is a reformulation based on the Gaussian Processes (GPs) \cite{carl2006gp}, which is a non-parametric Bayesian kernel method. Therefore, our model also can adapt its complexity flexibly to the complex data distributions in the real-world, without any heuristics or manual tuning of parameters.

Reformulating GPs for large-scale multi-task learning is non-trivial. To simplify calculations, we introduce a more efficient equivalent form of Kernel Fisher Discriminant Analysis (KFDA) to DGPLVM. Despite that the GaussianFace model can be optimized effectively using the Scaled Conjugate Gradient (SCG) technique, the inference is slow for large-scale data. We make use of GP approximations \cite{carl2006gp} and anchor graphs \cite{liu2010large} to speed up the process of inference and prediction, so as to scale our model to large-scale data. Our model can be applied to face verification in two different ways: as a binary classifier and as a feature extractor. In the former mode, given a pair of face images, we can directly compute the posterior likelihood for each class to make a prediction. In the latter mode, our model can automatically extract high-dimensional features for each pair of face images, and then feed them to a classifier to make the final decision.

The main contributions of this paper are as follows:
\begin{itemize}
  \item We propose a novel GaussianFace model for face verification by virtue of the multi-task learning constraint to DGPLVM. Our model can adapt to complex distributions, avoid over-fitting, exploit discriminative information, and take advantage of multiple source-domains data.
  \item We introduce a computationally more efficient equivalent form of KFDA to DGPLVM. This equivalent form reformulates KFDA to the kernel version consistent with the covariance function in GPs, which greatly simplifies calculations.
  \item We introduce approximation in GPs and anchor graphs to speed up the process of inference and prediction.
  \item We achieve superior performance on the challenging LFW benchmark \cite{LFWTech}, with an accuracy rate of 98.52\%, beyond human-level performance reported in \cite{kumar2009attribute}.
\end{itemize}

\section{Related Work}

Human and computer performance on face recognition has been compared extensively \cite{o2007face,o2012comparing,adler2007comparing,tang2004face,phillips2014comparison,bruce1998comparisons}. These studies have shown that computer-based algorithms were more accurate than humans in well-controlled environments (e.g., frontal view, natural expression, and controlled illumination), whilst still comparable to humans in the poor condition (e.g., frontal view, natural expression, and uncontrolled illumination). However, the above conclusion is only verified on face datasets with controlled variations, where only one factor changes at a time \cite{o2007face,o2012comparing}. To date, there has been virtually no work showing that computer-based algorithms could surpass human performance on unconstrained face datasets, such as LFW, which exhibits natural (multifactor) variations in pose, lighting, expression, race, ethnicity, age, gender, clothing, hairstyles, and other parameters.

There has been much work dealing with multifactor variations in face verification. For example, Simonyan \emph{et al.} applied the Fisher vector to face verification and achieved a good performance \cite{simonyan2004fisher}. However, the Fisher vector is derived from the Gaussian mixture model (GMM), where the number of Gaussians need to be specified by users, which means it cannot cover complex data automatically. Li \emph{et al.} proposed a non-parametric subspace analysis \cite{li2005nonparametric,li2009nonparametric}, but it is only a linear transformation and cannot cover the complex distributions. Besides, there also exist some approaches for utilizing plentiful source-domain data. Based on the Joint Bayesian algorithm \cite{chen2012bayesian}, Cao \emph{et al.} proposed a transfer learning approach \cite{Cao2013Transfer} by merging source-domain data with limited target-domain data. Since this transfer learning approach is based on the joint Bayesian model of original visual features, it is not suitable for handling the complex nonlinear data and the data with complex manifold structures. Moreover, the transfer learning approach in \cite{Cao2013Transfer} only considered two different domains, restricting its wider applications in large-scale data from multiple domains. More recently, Zhu \emph{et al.} \cite{ping2014} learned the transformation from face images under various poses and lighting conditions to a canonical view with a deep convolutional network. Sun \emph{et al.} \cite{sun2014} learned face representation with a deep model through face identification, which is a challenging multi-class prediction task. Taigman \emph{et al.} \cite{taigman2009multiple} first utilized explicit 3D face modeling to apply a piecewise affine transformation, and then derived a face representation from a nine-layer deep neural network. Although these methods have achieved high performances on LFW, many parameters of them must be determined in advance so that they are less flexible when dealing with complex data distributions.

The core of our algorithm is GPs. To the best of our knowledge, GPs methods and Multi-task learning with related GPs methods (MTGP) have not been applied for face verification. Actually, MTGP/GPs have been extensively studied in machine learning and computer vision in recent years \cite{bonilla2008multi,yu2005learning,chai2010multi,kim2006appearance,leen2011focused,rudovic2010coupled,skolidis2011bayesian,zhang2010multi,kim2007clustering}. However, most of them \cite{yu2005learning,chai2010multi,bonilla2008multi,rudovic2010coupled,kim2006appearance,skolidis2011bayesian,zhang2010multi} have only considered the symmetric multi-task learning, which means that all tasks have been assumed to be of equal importance, whereas our purpose is to enhance performance on a target task given all other source tasks. Leen \emph{et al.} proposed a MTGP model in the asymmetric setting \cite{leen2011focused} to focus on improving performance on the target task, and Kim \emph{et al.} developed a GP model for clustering \cite{kim2007clustering}, but their methods do not take the discriminative information of the covariance function into special account like DGPLVM. Although the discriminative information is considered in \cite{urtasun2007discriminative}, it does not apply multi-task learning to improve its performance. Salakhutdinov \emph{et al.} used a deep belief net to learn a good covariance kernel for GPs \cite{salakhutdinov2007using}. The limitation of such deep methods is that it is hard to determine which architecture for this network is optimal. Also, multi-task learning constraint was not considered in \cite{salakhutdinov2007using}.

\section{Preliminary}

In this section, we briefly review Gaussian Processes (GPs) for classification and clustering \cite{kim2007clustering}, and Gaussian Process Latent Variable Model (GPLVM) \cite{lawrence2003gaussian}. We use GPs method mainly due to the following three notable advantages. Firstly, as mentioned previously, it is a non-parametric method, which means it adapts its complexity flexibly to the complex data distributions in the real-world, without any heuristics or manual tuning of parameters. Secondly, GPs method can be computed effectively because of its closed-form marginal probability computation. Furthermore, its hyper-parameters can be learned from data automatically without using model selection methods such as cross validation, thereby avoiding the high computational cost. Thirdly, the inference of GPs is based on Bayesian rules, resulting in robustness to overfitting. We recommend Rasmussen and Williams's excellent monograph for further reading \cite{carl2006gp}.

\subsection{Gaussian Processes for Binary Classification}

Formally, for two-class classification, suppose that we have a training set $\mathcal{D}$ of $N$ observations, $\mathcal{D}=\{(\mathbf{x}_i, y_i)\}_{i=1}^N$, where the $i$-th input point $\mathbf{x}_i\in \mathbb{R}^D$ and its corresponding output $y_i$ is binary, with $y=1_i$ for one class and $y_i=-1$ for the other. Let $\mathbf{X}$ be the $N\times D$ matrix, where the row vectors represent all $n$ input points, and $\mathbf{y}$ be the column vector of all $n$ outputs. We define a latent variable $f_i$ for each input point $\mathbf{x}_i$, and let $\mathbf{f}=[f_1, \ldots, f_N]^{\top}$. A sigmoid function $\pi(\cdot)$ is imposed to squash the output of the latent function into $[0, 1]$, $\pi(f_i)=p(y_i=1|f_i)$. Assuming the data set is i.i.d, then the joint likelihood factorizes to \vspace{-2mm}
\begin{align}\label{equation:6}
p(\mathbf{y}|\mathbf{f})=\prod_{i=1}^{N}p(y_i|f_i)=\prod_{i=1}^{N}\pi(y_if_i).
\end{align}
Moreover, the posterior distribution over latent functions is \vspace{-2mm}
\begin{align}\label{equation:7}
p(\mathbf{f}|\mathbf{X},\mathbf{y},\boldsymbol{\theta})=\frac{p(\mathbf{y}|\mathbf{f})p(\mathbf{f}|\mathbf{X})}{p(\mathbf{y}|\mathbf{X},\boldsymbol{\theta})}.
\end{align}
Since neither $p(\mathbf{f}|\mathbf{X},\mathbf{y},\boldsymbol{\theta})$ nor $p(\mathbf{y}|\mathbf{f})$ can be computed analytically, the Laplace method is utilized to approximate the posterior \vspace{-1mm}
\begin{align}\label{equation:8}
p(\mathbf{f}|\mathbf{X},\mathbf{y},\boldsymbol{\theta})= \mathcal{N}(\hat{\mathbf{f}}, (\mathbf{K}^{-1}+\mathbf{W})^{-1}),
\end{align}
where $\hat{\mathbf{f}}=\arg\max_{\mathbf{f}}p(\mathbf{f}|\mathbf{X},\mathbf{y},\boldsymbol{\theta})$ and $\mathbf{W}=-\triangledown\triangledown\log p(\mathbf{f}|\mathbf{X},\mathbf{y},\boldsymbol{\theta})|_{{\mathbf{f}}=\hat{\mathbf{f}}}$. Then, we can obtain \vspace{-2mm}
\begin{align}\label{equation:9}
\log p(\mathbf{y}| \mathbf{X}, \boldsymbol{\theta})=
-\frac{1}{2}\hat{\mathbf{f}}^{\top}\mathbf{K}^{-1}\hat{\mathbf{f}}+\log p(\mathbf{y}|\hat{\mathbf{f}})-\frac{1}{2}\log |\mathbf{B}|.
\end{align}
where $|\mathbf{B}|=|\mathbf{K}|\cdot |\mathbf{K}^{-1}+\mathbf{W}|=|\mathbf{I}_n+\mathbf{W}^{\frac{1}{2}}\mathbf{K}\mathbf{W}^{\frac{1}{2}}|$. The optimal value of $\boldsymbol{\theta}$ can be acquired by using the gradient method to maximize Equation (\ref{equation:9}). Given any unseen test point $x_{\ast}$, the probability of its latent function $f_{\ast}$ is
\begin{align}\label{equation:10}
f_{\ast}|\mathbf{X},\mathbf{y},x_{\ast}\sim
\mathcal{N}(\mathbf{K}_{\ast}\mathbf{K}^{-1}\hat{\mathbf{f}}, \mathbf{K}_{\ast\ast}-\mathbf{K}_{\ast}\tilde{\mathbf{K}}^{-1}\mathbf{K}_{\ast}^{\top}),
\end{align}
where $\tilde{\mathbf{K}}=\mathbf{K}+\mathbf{W}^{-1}$. Finally, we squash $f_{\ast}$ to find the probability of class membership as follows
\begin{align}\label{equation:11}
\bar{\pi}(f_{\ast})=\int \pi(f_{\ast})p(f_{\ast}|\mathbf{X},\mathbf{y},x_{\ast})\text{d}f_{\ast}.
\end{align}

\subsection{Gaussian Processes for Clustering} \label{section:gpforclustering}

The principle of GP clustering is based on the key observation that the variances of predictive values are smaller in dense areas and larger in sparse areas. The variances can be employed as a good estimate of the support of a probability density function, where each separate support domain can be considered as a cluster. This observation can be explained from the variance function of any predictive data point $x_{\ast}$
\begin{align}\label{equation:14}
{\sigma}^2(x_{\ast})=\mathbf{K}_{\ast\ast}-\mathbf{K}_{\ast}\tilde{\mathbf{K}}^{-1}\mathbf{K}_{\ast}^{\top}.
\end{align}
If $x_{\ast}$ is in a sparse region, then $\mathbf{K}_{\ast}\tilde{\mathbf{K}}^{-1}\mathbf{K}_{\ast}^{\top}$ becomes small, which leads to large variance ${\sigma}^2(x_{\ast})$, and vice versa. Another good property of Equation (\ref{equation:14}) is that it does not depend on the labels, which means it can be applied to the unlabeled data.

To perform clustering, the following dynamic system associated with Equation (\ref{equation:14}) can be written as
\begin{align}\label{equation:15}
F(x)=-\triangledown{\sigma}^2(x).
\end{align}
The theorem in \cite{kim2007clustering} guarantees that almost all the trajectories approach one of the stable equilibrium points detected from Equation (\ref{equation:15}). After each data point finds its corresponding stable equilibrium point, we can employ a complete graph \cite{ben2002support,kim2007clustering} to assign cluster labels to data points with the stable equilibrium points. Obviously, the variance function in Equation (\ref{equation:14}) completely determines the performance of clustering.

\subsection{Gaussian Process Latent Variable Model}

Let $\textbf{Z}=[\textbf{z}_1, \ldots, \textbf{z}_N]^{\top}$ denote the matrix whose rows represent corresponding positions of $\textbf{X}$ in latent space, where $\textbf{z}_i \in \mathbb{R}^d$ ($d \ll D$). The Gaussian Process Latent Variable Model (GPLVM) can be interpreted as a Gaussian process mapping from a low dimensional latent space to a high dimensional data set, where the locale of the points in latent space is determined by maximizing the Gaussian process likelihood with respect to $\textbf{Z}$. Given a covariance function for the Gaussian process, denoted by $k(\cdot, \cdot)$, the likelihood of the data given the latent positions is as follows,
\begin{align} \label{equation:gplvm_1}
p(\textbf{X}|\textbf{Z}, \boldsymbol{\theta})=\frac{1}{\sqrt{(2\pi)^{ND}|\textbf{K}|^{D}}}\exp
\Big(-\frac{1}{2}\text{tr}(\textbf{K}^{-1}\textbf{X}\textbf{X}^{\top})\Big),
\end{align}
where $\textbf{K}_{i,j}=k(\textbf{z}_i, \textbf{z}_j)$. Therefore, the posterior can be written as
\begin{align} \label{equation:gplvm_likelihood}
p(\textbf{Z}, \boldsymbol{\theta}|\textbf{X})=\frac{1}{\mathcal{Z}_a}
p(\textbf{X}|\textbf{Z}, \boldsymbol{\theta})p(\textbf{Z})p(\boldsymbol{\theta}),
\end{align}
where $\mathcal{Z}_a$ is a normalization constant, the uninformative priors over $\boldsymbol{\theta}$, and the simple spherical Gaussian priors over $\textbf{Z}$ are introduced \cite{urtasun2007discriminative}. To obtain the optimal $\boldsymbol{\theta}$ and $\textbf{Z}$, we need to optimize the above likelihood (\ref{equation:gplvm_likelihood}) with respect to $\boldsymbol{\theta}$ and $\textbf{Z}$, respectively.

\section{GaussianFace}

In order to automatically learn discriminative features or covariance function, and to take advantage of source-domain data to improve the performance in face verification, we develop a principled GaussianFace model by including the multi-task learning constraint into Discriminative Gaussian Process Latent Variable Model (DGPLVM) \cite{urtasun2007discriminative}.

\subsection{DGPLVM Reformulation}

The DGPLVM is an extension of GPLVM, where the discriminative prior is placed over the latent positions, rather than a simple spherical Gaussian prior. The DGPLVM uses the discriminative prior to encourage latent positions of the same class to be close and those of different classes to be far. Since face verification is a binary classification problem and the GPs mainly depend on the kernel function, it is natural to use Kernel Fisher Discriminant Analysis (KFDA) \cite{kim2006optimal} to model class structures in kernel spaces. For simplicity of inference in the followings, we introduce another equivalent formulation of KFDA to replace the one in \cite{urtasun2007discriminative}.

KFDA is a kernelized version of linear discriminant analysis method. It finds the direction defined by a kernel in a feature space, onto which the projections of positive and negative classes are well separated by maximizing the ratio of the between-class variance to the within-class variance. Formally, let $\{\textbf{z}_1,\ldots,\textbf{z}_{N_+}\}$ denote the positive class and $\{\textbf{z}_{N_++1},\ldots,\textbf{z}_N\}$ the negative class, where the numbers of positive and negative classes are $N_+$ and $N_-=N-N_+$, respectively. Let $\mathbf{K}$ be the kernel matrix. Therefore, in the feature space, the two sets $\{\phi_{\mathbf{K}}(\textbf{z}_1),\ldots,\phi_{\mathbf{K}}(\textbf{z}_{N_+})\}$ and $\{\phi_{\mathbf{K}}(\textbf{z}_{N_++1}),\ldots,\phi_{\mathbf{K}}(\textbf{z}_N)\}$ represent the positive class and the negative class, respectively. The optimization criterion of KFDA is to maximize the ratio of the between-class variance to the within-class variance
\begin{align} \label{equation:kfda_j}
J(\omega, \mathbf{K})=\frac{(\mathbf{w}^{\top}(\mathbf{\mu}_{\mathbf{K}}^+-\mathbf{\mu}_{\mathbf{K}}^-))^2}{\mathbf{w}^{\top}(\mathbf{\Sigma}_{\mathbf{K}}^++\mathbf{\Sigma}_{\mathbf{K}}^-+\lambda\mathbf{I}_N)\mathbf{w}},
\end{align}
where $\lambda$ is a positive regularization parameter, $\mathbf{\mu}_{\mathbf{K}}^+=\frac{1}{N_+}\sum_{i=1}^{N_+}\phi_{\mathbf{K}}(\textbf{z}_i)$, $\mathbf{\mu}_{\mathbf{K}}^-=\frac{1}{N_-}\sum_{i=N_++1}^{N}\phi_{\mathbf{K}}(\textbf{z}_i)$, $\mathbf{\Sigma}_{\mathbf{K}}^+=\frac{1}{N_+}\sum_{i=1}^{N_+}(\phi_{\mathbf{K}}(\textbf{z}_i)-\mathbf{\mu}_{\mathbf{K}}^+)(\phi_{\mathbf{K}}(\textbf{z}_i)-\mathbf{\mu}_{\mathbf{K}}^+)^{\top}$, and $\mathbf{\Sigma}_{\mathbf{K}}^-=\frac{1}{N_-}\sum_{i=N_++1}^{N}(\phi_{\mathbf{K}}(\textbf{z}_i)-\mathbf{\mu}_{\mathbf{K}}^-)(\phi_{\mathbf{K}}(\textbf{z}_i)-\mathbf{\mu}_{\mathbf{K}}^-)^{\top}$.

In this paper, however, we focus on the covariance function rather than the latent positions. To simplify calculations, we represent Equation (\ref{equation:kfda_j}) with the kernel function, and let the kernel function have the same form as the covariance function. Therefore, it is natural to introduce a more efficient equivalent form of KFDA with certain assumptions as Kim \emph{et al.} points out \cite{kim2006optimal}, i.e., maximizing Equation (\ref{equation:kfda_j}) is equivalent to maximizing the following equation
\begin{align}\label{equation:kfda_jn}
J^{\ast}=
\frac{1}{\lambda}\big(\mathbf{a}^{\top}\mathbf{K}\mathbf{a}-\mathbf{a}^{\top}\mathbf{K}\mathbf{A}(\lambda\mathbf{I}_n+\mathbf{A}\mathbf{K}\mathbf{A})^{-1}\mathbf{A}\mathbf{K}\mathbf{a}\big),
\end{align}
where
\begin{align}
\mathbf{a}=&[\frac{1}{n_+}\mathbf{1}_{N_+}^{\top}, -\frac{1}{N_-}\mathbf{1}_{N_-}^{\top}] \nonumber\\  \mathbf{A}=&\text{diag}\Big(\frac{1}{\sqrt{N_+}}\big(\mathbf{I}_{N_+}-\frac{1}{N_+}\mathbf{1}_{N_+}\mathbf{1}_{N_+}^{\top}\big),\nonumber\\ &\quad \quad\frac{1}{\sqrt{N_-}}\big(\mathbf{I}_{N_-}-\frac{1}{N_-}\mathbf{1}_{N_-}\mathbf{1}_{N_-}^{\top}\big)\Big). \nonumber
\end{align}
Here, $\mathbf{I}_N$ denotes the $N\times N$ identity matrix and $\mathbf{1}_N$ denotes the length-$N$ vector of all ones in $\mathbb{R}^N$.

Therefore, the discriminative prior over the latent positions in DGPLVM can be written as
\begin{align}\label{equation:discriminative}
p(\textbf{Z})=\frac{1}{\mathcal{Z}_b}\exp\Big(-\frac{1}{\sigma^2}J^{\ast}\Big),
\end{align}
where $\mathcal{Z}_b$ is a normalization constant, and $\sigma^2$ represents a global scaling of the prior.

The covariance matrix obtained by DGPLVM is discriminative and more flexible than the one used in conventional GPs for classification (GPC), since they are learned based on a discriminative criterion, and more degrees of freedom are estimated than conventional kernel hyper-parameters.

\subsection{Multi-task Learning Constraint}

From an asymmetric multi-task learning perspective, the tasks should be allowed to share common hyper-parameters of the covariance function. Moreover, from an information theory perspective, the information cost between target task and multiple source tasks should be minimized. A natural way to quantify the information cost is to use the mutual entropy, because it is the measure of the mutual dependence of two distributions. For multi-task learning, we extend the mutual entropy to multiple distributions as follows
\begin{align}\label{equation:multitast}
\mathcal{M}=H(p_t)-\frac{1}{S}\sum_{i=1}^{S}H(p_t|p_i),
\end{align}
where $H(\cdot)$ is the marginal entropy, $H(\cdot|\cdot)$ is the conditional entropy, $S$ is the number of source tasks, $\{p_i\}_{i=1}^S$, and $p_t$ are the probability distributions of source tasks and target task, respectively.

\subsection{GaussianFace Model}

In this section, we describe our GaussianFace model in detail. Suppose we have $S$ source-domain datasets $\{\mathbf{X}_1,\ldots,\mathbf{X}_S\}$ and a target-domain data $\mathbf{X}_T$. For each source-domain data or target-domain data $\mathbf{X}_i$, according to Equation (\ref{equation:gplvm_1}), we write its marginal likelihood
\begin{align}\label{equation:marginal}
p(\textbf{X}_i|\textbf{Z}_i, \boldsymbol{\theta})=\frac{1}{\sqrt{(2\pi)^{ND}|\textbf{K}|^{D}}}\exp
\Big(-\frac{1}{2}\text{tr}(\textbf{K}^{-1}\textbf{X}_i\textbf{X}_i^{\top})\Big).
\end{align}
where $\textbf{Z}_i$ represents the domain-relevant latent space. For each source-domain data and target-domain data, their covariance functions $\mathbf{K}$ have the same form because they share the same hyper-parameters $\boldsymbol{\theta}$. In this paper, we use a widely used kernel
\begin{align}
\mathbf{K}_{i,j}=k_{\boldsymbol{\theta}}(\mathbf{z}_i,\mathbf{z}_j)
=&\theta_0\exp\Big(-\frac{1}{2}\sum_{m=1}^{d}\theta_m(\mathbf{z}_i^m-\mathbf{z}_j^m)^2\Big)\nonumber \\
&+\theta_{d+1}+\frac{\delta_{\mathbf{z}_i,\mathbf{z}_j}}{\theta_{d+2}},
\end{align}
where $\boldsymbol{\theta}=\{\theta_i\}_{i=0}^{d+2}$ and $d$ is the dimension of the data point. Then, from Equations (\ref{equation:gplvm_likelihood}), learning the DGPLVM is equivalent to optimizing
\begin{align}\label{equation:discriminative_constraint}
p(\textbf{Z}_i, \boldsymbol{\theta}|\textbf{X}_i)=\frac{1}{\mathcal{Z}_a}
p(\textbf{X}_i|\textbf{Z}_i, \boldsymbol{\theta})p(\textbf{Z}_i)p(\boldsymbol{\theta}),
\end{align}
where $p(\textbf{X}_i|\textbf{Z}_i, \boldsymbol{\theta})$ and $p(\textbf{Z}_i)$ are respectively represented in (\ref{equation:marginal}) and (\ref{equation:discriminative}). According to the multi-task learning constraint in Equation (\ref{equation:multitast}), we can attain
\begin{align}\label{equation:multitast_constraint}
\mathcal{M}=&H(p(\textbf{Z}_T, \boldsymbol{\theta}|\textbf{X}_T))\nonumber \\
&-\frac{1}{S}\sum_{i=1}^{S}H(p(\textbf{Z}_T, \boldsymbol{\theta}|\textbf{X}_T)|p(\textbf{Z}_i, \boldsymbol{\theta}|\textbf{X}_i)).
\end{align}
From Equations (\ref{equation:marginal}), (\ref{equation:discriminative_constraint}), and (\ref{equation:multitast_constraint}), we know that learning the GaussianFace model amounts to minimizing the following marginal likelihood
\begin{align}\label{equation:MTL-DGPLVM}
\mathcal{L}_{Model}=-\log p(\textbf{Z}_T, \boldsymbol{\theta}|\textbf{X}_T)-\beta \mathcal{M},
\end{align}
where the parameter $\beta$ balances the relative importance between the target-domain data and the multi-task learning constraint.

\subsection{Optimization}

For the model optimization, we first expand Equation (\ref{equation:MTL-DGPLVM}) to obtain the following equation (ignoring the constant items)
\begin{align}\label{equation:expasion}
\mathcal{L}_{Model}=&-\log P_T+\beta P_T \log P_T \nonumber \\
&+\frac{\beta}{S}\sum_{i=1}^{S}\big(P_{T, i} \log P_i-P_{T, i} \log P_{T, i}\big),
\end{align}
where $P_i=p(\textbf{Z}_i, \boldsymbol{\theta}|\textbf{X}_i)$ and $P_{i, j}$ means that its corresponding covariance function is computed on both $\mathbf{X}_i$ and $\mathbf{X}_j$. We can now optimize Equation (\ref{equation:expasion}) with respect to the hyper-parameters $\boldsymbol{\theta}$ and the latent positions $\textbf{Z}_i$ by the Scaled Conjugate Gradient (SCG) technique. Since we focus on the covariance matrix in this paper, here we only present the derivations of hyper-parameters. It is easy to get
\begin{align}
\frac{\partial \mathcal{L}_{Model}}{\partial \theta_j}=&\Big(\beta(\log P_T+1)-\frac{1}{P_T}\Big)\frac{\partial P_T}{\partial \theta_j} \nonumber \\
&+\frac{\beta}{S}\sum_{i=1}^{S}\frac{P_{T,i}}{P_i}\cdot\frac{\partial P_i}{\partial \theta_j}\nonumber \\
&+\frac{\beta}{S}\sum_{i=1}^{S}(\log P_i-\log P_{T, i}-1)\frac{\partial P_{T, i}}{\partial \theta_j}. \nonumber
\end{align}
The above equation depends on the form $\frac{\partial P_i}{\partial \theta_j}$ as follows (ignoring the constant items)
\begin{align}
\frac{\partial P_i}{\partial \theta_j}=&P_i\frac{\partial \log P_i}{\partial \theta_j}\nonumber \\
\approx & P_i\Big(\frac{\partial \log p(\textbf{X}_i|\textbf{Z}_i, \boldsymbol{\theta})}{\partial \theta_j} + \frac{\partial \log p(\textbf{Z}_i)}{\partial \theta_j}+\frac{\partial \log p(\boldsymbol{\theta})}{\partial \theta_j}\Big).\nonumber
\end{align}
The above three terms can be easily obtained (ignoring the constant items) by
\begin{align}
\frac{\partial \log p(\textbf{X}_i|\textbf{Z}_i, \boldsymbol{\theta})}{\partial \theta_j}
\approx &-\frac{D}{2}\text{Tr}\Big(\mathbf{K}^{-1}\frac{\partial \mathbf{K}}{\partial \theta_j}\Big)\nonumber\\&+\frac{1}{2}\text{Tr}\Big(\mathbf{K}^{-1}\textbf{X}_i\textbf{X}_i^{\top}\mathbf{K}^{-1}\frac{\partial \mathbf{K}}{\partial \theta_j}\Big)
,\nonumber \\
\frac{\partial \log p(\textbf{Z}_i)}{\partial \theta_j}\approx&-\frac{1}{\sigma^2}\frac{\partial J_i^{\ast}}{\partial \theta_j}\nonumber \\
=&-\frac{1}{\lambda\sigma^2}\Big(\mathbf{a}^{\top}\frac{\partial \mathbf{K}}{\partial \theta_j}\mathbf{a}-\mathbf{a}^{\top}\frac{\partial \mathbf{K}}{\partial \theta_j}\tilde{\mathbf{A}}\mathbf{a}\nonumber \\
&+\mathbf{a}^{\top}\mathbf{K}\tilde{\mathbf{A}}\frac{\partial \mathbf{K}}{\partial \theta_j}\tilde{\mathbf{A}}\mathbf{K}\mathbf{a}-\mathbf{a}^{\top}\mathbf{K}\tilde{\mathbf{A}}\frac{\partial \mathbf{K}}{\partial \theta_j}\mathbf{a}\Big), \nonumber \\
\frac{\partial \log p(\boldsymbol{\theta})}{\partial \theta_j}=& \frac{1}{\theta_j},\nonumber
\end{align}
where $\tilde{\mathbf{A}}=\mathbf{A}(\lambda \mathbf{I}_n+\mathbf{A}\mathbf{K}\mathbf{A})^{-1}\mathbf{A}$. Thus, the desired derivatives have been obtained.

\subsection{Speedup} \label{section:speedup}

In the GaussianFace model, we need to invert the large matrix when doing inference and prediction. For large problems, both storing the matrix and solving the associated linear systems are computationally prohibitive. In this paper, we use the anchor graphs method \cite{liu2010large} to speed up this process. To put it simply, we first select $q$ ($q \ll n$) anchors to cover a cloud of $n$ data points, and form an $n\times q$ matrix $\mathbf{Q}$, where $\mathbf{Q}_{i,j}=k_{\boldsymbol{\theta}}(\mathbf{z}_i, \mathbf{z}_j)$. $\mathbf{z}_i$ and $\mathbf{z}_j$ are from $n$ latent data points and $q$ anchors, respectively. Then the original kernel matrix $\mathbf{K}$ can be approximated as $\mathbf{K}\approx\mathbf{Q}\mathbf{Q}^{\top}$. Using the Woodbury identity \cite{higham1996accuracy}, computing the $n\times n$ matrix $\mathbf{Q}\mathbf{Q}^{\top}$ can be transformed into computing the $q\times q$ matrix $\mathbf{Q}^{\top}\mathbf{Q}$, which is more efficient.\\

\textbf{Speedup on Inference}  When optimizing Equation (\ref{equation:MTL-DGPLVM}), we need to invert the matrix $(\lambda\mathbf{I}_n+\mathbf{A}\mathbf{K}\mathbf{A})$. During inference, we take $q$ k-means clustering centers as anchors to form $\textbf{Q}$. Substituting $\mathbf{K}\approx\mathbf{Q}\mathbf{Q}^{\top}$ into $(\lambda\mathbf{I}_n+\mathbf{A}\mathbf{K}\mathbf{A})$, and then using the Woodbury identity, we get
\begin{align}
(\lambda\mathbf{I}_n+\mathbf{A}\mathbf{K}\mathbf{A})^{-1}\approx(\lambda\mathbf{I}_n+\mathbf{A}\mathbf{Q}\mathbf{Q}^{\top}\mathbf{A})^{-1} ~~~~~~~~~~~~~~~~~~~~~~~~~\nonumber \\
=\lambda^{-1}\mathbf{I}_n-\lambda^{-1}\mathbf{A}\mathbf{Q}(\lambda\mathbf{I}_q+\mathbf{Q}^{\top}\mathbf{A}\mathbf{A}\mathbf{Q})^{-1}\mathbf{Q}^{\top}\mathbf{A}.
\nonumber
\end{align}
Similarly, let $\mathbf{K}^{-1} \approx (\mathbf{K}+\tau \mathbf{I})^{-1}$ where $\tau$ a constant term, then we can get
\begin{align}
\mathbf{K}^{-1} \approx (\mathbf{K}+\tau \mathbf{I})^{-1} \approx
\tau^{-1}\mathbf{I}_n-\tau^{-1}\mathbf{Q}(\tau\mathbf{I}_q+\mathbf{Q}^{\top}\mathbf{Q})^{-1}\mathbf{Q}^{\top}.\nonumber
\end{align}

\textbf{Speedup on Prediction}  When we compute the predictive variance $\sigma(\mathbf{z}_{\ast})$, we need to invert the matrix $(\mathbf{K}+\mathbf{W}^{-1})$. At this time, we can use the method in Section \ref{section:gpforclustering} to calculate the accurate clustering centers that can be regarded as the anchors. Using the Woodbury identity again, we obtain
\begin{align}
(\mathbf{K}+\mathbf{W}^{-1})^{-1}\approx\mathbf{W}-\mathbf{W}\mathbf{Q}(\mathbf{I}_q+\mathbf{Q}^{\top}\mathbf{W}\mathbf{Q})^{-1}\mathbf{Q}^{\top}\mathbf{W}, \nonumber
\end{align}
where $(\mathbf{I}_q+\mathbf{Q}^{\top}\mathbf{W}\mathbf{Q})$ is only a $q\times q$ matrix, and its inverse matrix can be computed more efficiently.

\section{GaussianFace Model for Face Verification} \label{section:model}

In this section, we describe two applications of the GaussianFace model to face verification: as a binary classifier and as a feature extractor.

Each face image is first normalized to $150\times120$ size by an affine transformation based on five landmarks (two eyes, nose, and two mouth corners). The image is then divided into overlapped patches of $25\times25$ pixels with a stride of $2$ pixels. Each patch within the image is mapped to a vector by a certain descriptor, and the vector is regarded as the feature of the patch, denoted by $\{\mathbf{x}_p^A\}_{p=1}^{P}$ where $P$ is the number of patches within the face image $A$. In this paper, the multi-scale LBP feature of each patch is extracted \cite{Dong2013Blessing}. The difference is that the multi-scale LBP descriptors are extracted at the center of each patch instead of accurate landmarks.

\subsection{GaussianFace Model as a Binary Classifier}

For classification, our model can be regarded as an approach to learn a covariance function for GPC, as shown in Figure \ref{fig:pipelines} (a). Here, for a pair of face images $A$ and $B$ from the same (or different) person, let the similarity vector $\mathbf{x}_i=[s_1, \ldots, s_p, \ldots, s_P]^{\top}$ be the input data point of the GaussianFace model, where $s_p$ is the similarity of $\mathbf{x}_p^A$ and $\mathbf{x}_p^B$, and its corresponding output is $y_i=1$ (or $-1$). With the learned hyper-parameters of covariance function from the training data, given any un-seen pair of face images, we first compute its similarity vector $\mathbf{x}_{\ast}$ using the above method, then estimate its latent representation $\mathbf{z}_{\ast}$ using the same method in \cite{urtasun2007discriminative}, and finally predict whether the pair is from the same person through Equation (\ref{equation:11}). In this paper, we prescribe the sigmoid function $\pi(\cdot)$ to be the cumulative Gaussian distribution $\Phi(\cdot)$, which can be solved analytically as $
\bar{\pi}_{\ast}=\Phi\Big(\frac{\bar{f}_{\ast}(\mathbf{z}_{\ast})}{\sqrt{1+\sigma^2(\mathbf{z}_{\ast})}}\Big)$, where $\sigma^2(\mathbf{z}_{\ast})=\mathbf{K}_{\ast\ast}-\mathbf{K}_{\ast}\tilde{\mathbf{K}}^{-1}\mathbf{K}_{\ast}^{\top}$ and $\bar{f}_{\ast}(\mathbf{z}_{\ast})=\mathbf{K}_{\ast}\mathbf{K}^{-1}\hat{\mathbf{f}}$ from Equation (\ref{equation:10}) \cite{carl2006gp}. We call the method \emph{GaussianFace-BC}.

\subsection{GaussianFace Model as a Feature Extractor}

\begin{figure}[t]
  \centering
  \includegraphics[width=1.05\linewidth]{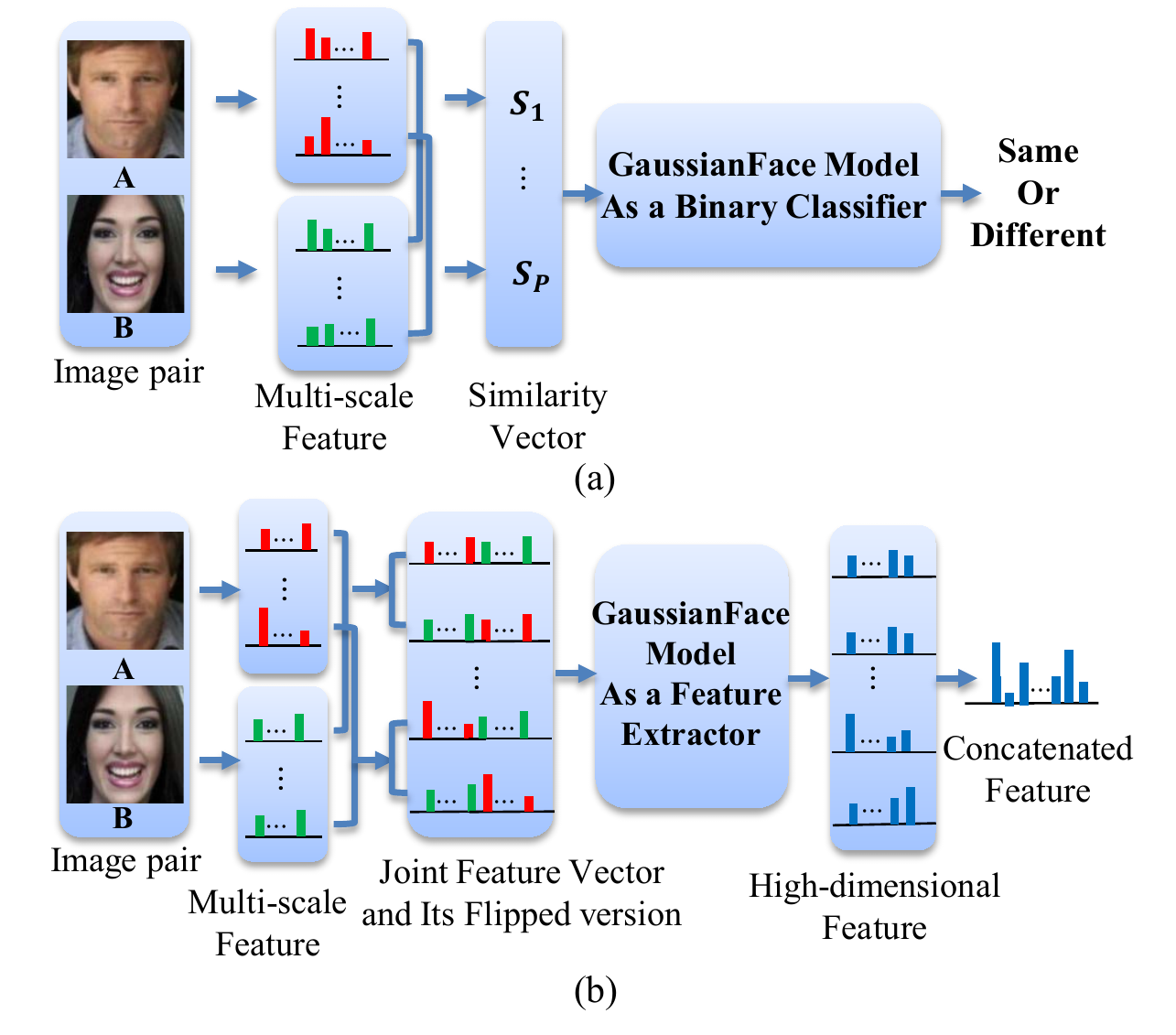}
  \caption{Two approaches based on GaussianFace model for face verification. (a) GaussianFace model as a binary classifier. (b) GaussianFace model as a feature extractor.
  }\label{fig:pipelines}
\end{figure}

As a feature extractor, our model can be regarded as an approach to automatically extract facial features, shown in Figure \ref{fig:pipelines} (b). Here, for a pair of face images $A$ and $B$ from the same (or different) person, we regard the joint feature vector $\mathbf{x}_i=[(\mathbf{x}_i^A)^{\top}, (\mathbf{x}_i^B)^{\top}]^{\top}$ as the input data point of the GaussianFace model, and its corresponding output is $y_i=1$ (or $-1$). To enhance the robustness of our approach, the flipped form of $\mathbf{x}_i$ is also included; for example, $\mathbf{x}_i=[(\mathbf{x}_i^B)^{\top}, (\mathbf{x}_i^A)^{\top}]^{\top}$. After the hyper-parameters of covariance function are learnt from the training data, we first estimate the latent representations of the training data using the same method in \cite{urtasun2007discriminative}, then can use the method in Section \ref{section:gpforclustering} to group the latent data points into different clusters automatically. Suppose that we finally obtain $C$ clusters. The centers of these clusters are denoted by $\{\mathbf{c}_i\}_{i=1}^{C}$, the variances of these clusters by $\{{\Sigma}_i^2\}_{i=1}^{C}$, and their weights by $\{w_i\}_{i=1}^{C}$ where $w_i$ is the ratio of the number of latent data points from the $i$-th cluster to the number of all latent data points. Then we refer to each $\mathbf{c}_i$ as the input of Equation (\ref{equation:10}), and we can obtain its corresponding probability $p_i$ and variance ${\sigma}_i^{2}$. In fact, $\{\mathbf{c}_i\}_{i=1}^{C}$ can be regarded as a codebook generated by our model.

For any un-seen pair of face images, we also first compute its joint feature vector $\mathbf{x}_{\ast}$ for each pair of patches, and estimate its latent representation $\mathbf{z}_{\ast}$. Then we compute its first-order and second-order statistics to the centers. Similarly, we regard $\mathbf{z}_{\ast}$ as the input of Equation (\ref{equation:10}), and can also obtain its corresponding probability $p_{\ast}$ and variance $\sigma_{\ast}^2$. The statistics and variance of $\mathbf{z}_{\ast}$ are represented as its high-dimensional facial features, denoted by $\hat{\mathbf{z}}_{\ast}=[{\Delta}_1^{1}, {\Delta}_1^{2}, {\Delta}_1^{3}, {\Delta}_1^{4}, \ldots, {\Delta}_C^{1}, {\Delta}_C^{2}, {\Delta}_C^{3}, {\Delta}_C^{4}]^{\top}$, where ${\Delta}_i^{1}=w_i\Big(\frac{\mathbf{z}_{\ast}-\mathbf{c}_i}{\Sigma_i}\Big)$, ${\Delta}_i^{2}=w_i\Big(\frac{\mathbf{z}_{\ast}-\mathbf{c}_i}{\Sigma_i}\Big)^2$, ${\Delta}_i^{3}=\log \frac{p_{\ast}(1- p_i)}{p_i(1-p_{\ast})}$, and ${\Delta}_i^{4}=\frac{\sigma_{\ast}^2}{{\sigma}_i^{2}}$. We then concatenate all of the new high-dimensional features from each pair of patches to form the final new high-dimensional feature for the pair of face images. The new high-dimensional facial features not only describe how the distribution of features of an un-seen face image differs from the distribution fitted to the features of all training images, but also encode the predictive information including the probabilities of label and uncertainty. We call this approach \emph{GaussianFace-FE}.

\section{Experimental Settings}

\begin{figure}[t]
  \centering
  \includegraphics[width=\linewidth]{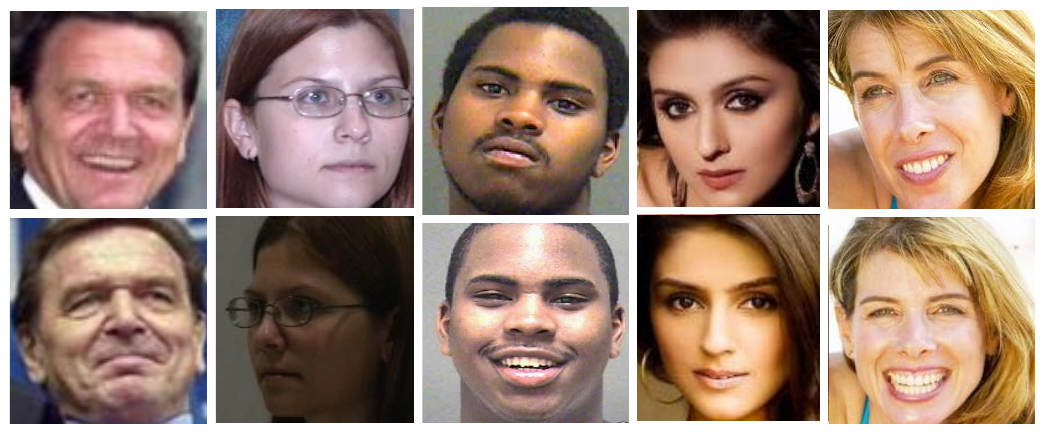}
  \caption{Samples of the datasets in our experiments. From left to right: LFW, Multi-PIE, MORPH, Web Images, and Life Photos.
  }\label{fig:samples}
\end{figure}

In this section, we conduct experiments on face verification. We start by introducing the source-domain datasets and the target-domain dataset in all of our experiments (see Figure \ref{fig:samples} for examples). The source-domain datasets include four different types of datasets as follows:\\
\textbf{Multi-PIE} \cite{gross2010multi}. This dataset contains face images from 337 subjects under 15 view points and 19 illumination conditions in four recording sessions. These images are collected under controlled conditions.\\
\textbf{MORPH} \cite{ricanek2006morph}. The MORPH database contains 55,000 images of more than 13,000 people within the age ranges of 16 to 77. There are an average of 4 images per individual.\\
\textbf{Web Images}\footnote[2]{These two datasets are collected by our own from the Web. It is guaranteed that these two datasets are mutually exclusive with the LFW dataset.}. This dataset contains around 40,000 facial images from 3261 subjects; that is, approximately 10 images for each person. The images were collected from the Web with significant variations in pose, expression, and illumination conditions.\\
$\textbf{Life Photos}^{\textcolor{black}{\text{2}}}$. This dataset contains approximately 5000 images of 400 subjects collected online. Each subject has roughly 10 images.

If not otherwise specified, the target-domain dataset is the benchmark of face
verification as follows:\\
\textbf{LFW} \cite{LFWTech}. This dataset contains 13,233 uncontrolled face images of 5749 public figures with variety of pose, lighting, expression, race, ethnicity, age, gender, clothing, hairstyles, and other parameters. All of these images are collected from the Web.

We use the LFW dataset as the target-domain dataset because it is well known as a challenging benchmark. Using it also allows us to compare directly with other existing face verification methods \cite{Cao2013Transfer,berg2012tom,Dong2013Blessing,simonyan2004fisher,chen2012bayesian,yin2011associate,visionlab,liaurora,fan2014learning}. Besides, this dataset provides a large set of relatively unconstrained face images with complex variations as described above, and has proven difficult for automatic face verification methods \cite{LFWTech,kumar2009attribute}. In all the experiments conducted on LFW, we strictly follow the standard unrestricted protocol of LFW \cite{LFWTech}. More precisely, during the training
procedure, the four source-domain datasets are: Web Images, Multi-PIE, MORPH, and Life Photos, the target-domain dataset is the training set in View 1 of LFW, and the validation set is the test set in View 1 of LFW. At the test time, we follow the standard 10-fold cross-validation protocol to test our model in View 2 of LFW.

For each one of the four source-domain datasets, we randomly sample 20,000 pairs of matched images and 20,000 pairs of mismatched images. The training partition and the testing partition in all of our experiments are mutually exclusive. In other words, there is no identity overlap among the two partitions.

For the experiments below, ``The Number of SD'' means ``the Number of Source-Domain datasets that are fed into the GaussianFace model for training''. By parity of reasoning, if ``The Number of SD'' is $i$, that means the first $i$ source-domain datasets are used for model training. Therefore, if ``The Number of SD'' is 0, models are trained with the training data from target-domain data only.\\

\textbf{Implementation details}. Our model involves four important parameters: $\lambda$ in (\ref{equation:kfda_jn}), $\sigma$ in (\ref{equation:discriminative}), $\beta$ in (\ref{equation:MTL-DGPLVM}), and the number of anchors $q$ in \emph{Speedup on Inference} \footnote[3]{The other parameters, such as the hyper-parameters in the kernel function and the number of anchors in \emph{Speedup on Prediction}, can be automatically learned from the data.}. Following the same setting in \cite{kim2006optimal}, the regularization parameter $\lambda$ in (\ref{equation:kfda_jn}) is fixed to $10^{-8}$. $\sigma$ reflects the tradeoff between our method's ability to discriminate (small $\sigma$) and its ability to generalize (large $\sigma$), and $\beta$ balances the relative importance between the target-domain data and the multi-task learning constraint. Therefore, the validation set (the test set in View 1 of LFW) is used for selecting $\sigma$ and $\beta$. Each time we use different number of source-domain datasets for training, the corresponding optimal $\sigma$ and $\beta$ should be selected on the validation set.

Since we collected a large number of image pairs for training (20,000 matched pairs and 20,000 mismatched pairs from each source-domain dataset), and our model is based on the kernel method, thus an important consideration is how to efficiently approximate the kernel matrix using a low-rank method in the limited space and time. We adopt the anchor graphs method (see Section \ref{section:speedup}) for kernel approximation. In our experiments, we take two steps to determine the number of anchor points. In the first step, the optimal $\sigma$ and $\beta$ are selected on the validation set in each experiment. In the second step, we fix $\sigma$ and $\beta$, and then tune the number of anchor points. We vary the number of anchor points to train our model on the training set, and test it on the validation set. We report the average accuracy for our model over 10 trials. After we consider the trade-off between memory and running time in practice, the number of anchor points with the best average accuracy is determined in each experiments.

\section{Experimental Results}

In this section, we conduct five experiments to demonstrate the validity of the GaussianFace model.

\subsection{Comparisons with Other MTGP/GP Methods}

\begin{figure*}
  \centering
  \includegraphics[width=0.95\linewidth]{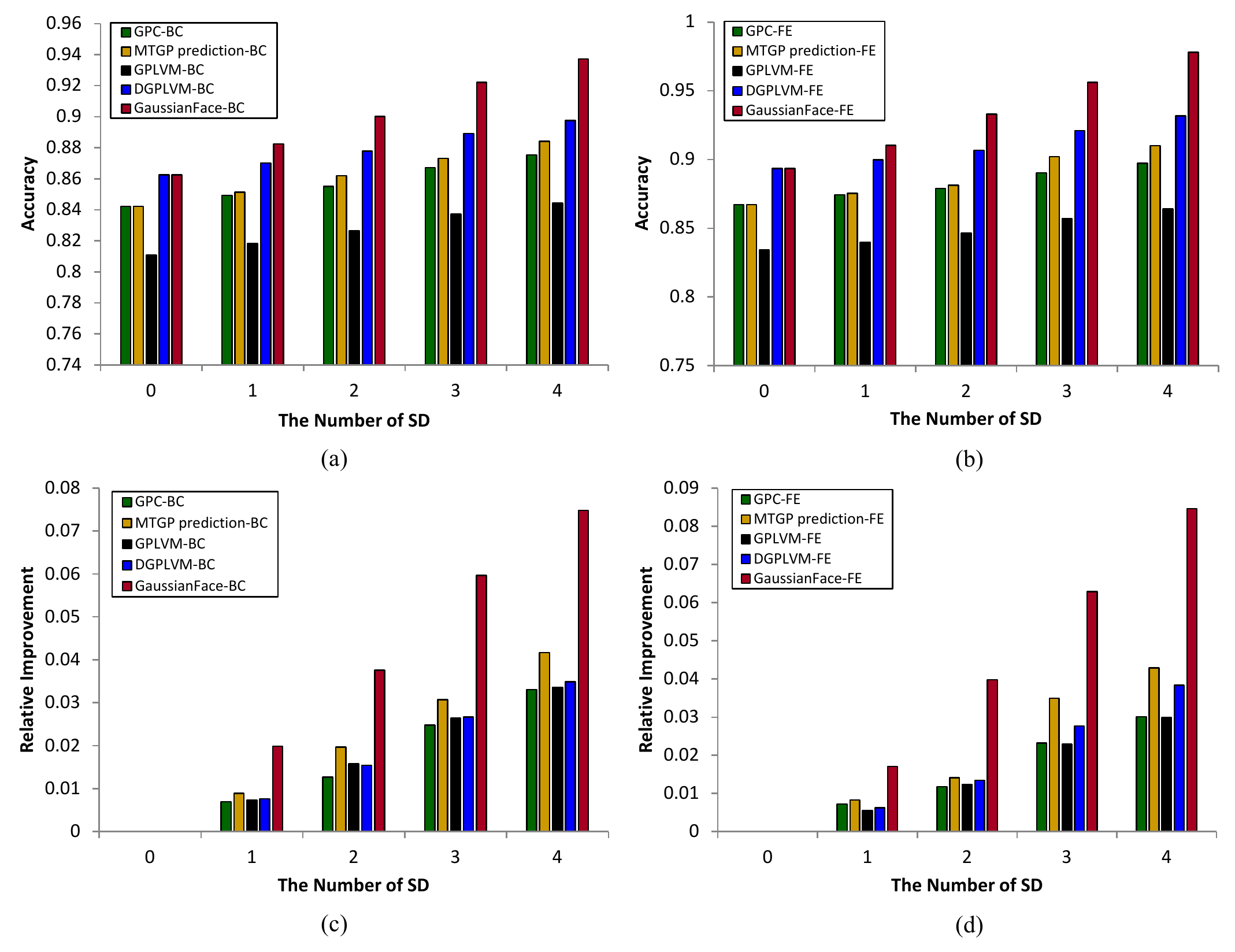}
  \caption{(a) The accuracy rate (\%) of the GaussianFace-BC model and other competing MTGP/GP methods as a binary classifier. (b) The accuracy rate (\%) of the GaussianFace-FE model and other competing MTGP/GP methods as a feature extractor. (c) The relative improvement of each method as a binary classifier with the increasing number of SD, compared to their performance when the number of SD is 0. (d) The relative improvement of each method as a feature extractor with the increasing number of SD, compared to their performance when the number of SD is 0.
  }\label{fig:comparisonwithGPs}
\end{figure*}

Since our model is based on GPs, it is natural to compare our model with four popular GP models: GPC \cite{carl2006gp}, MTGP prediction \cite{bonilla2008multi}, GPLVM \cite{lawrence2003gaussian}, and DGPLVM \cite{urtasun2007discriminative}. For fair comparisons, all these models are trained on multiple source-domain datasets using the same two methods as our GaussianFace model described in Section \ref{section:model}. After the hyper-parameters of covariance function are learnt for each model, we can regard each model as a binary classifier and a feature extractor like ours, respectively. Figure \ref{fig:comparisonwithGPs} shows that our model significantly outperforms the other four GPs models, and the superiority of our model becomes more obvious as the number of source-domain datasets increases.

\subsection{Comparisons with Other Binary Classifiers} \label{section:bc}

Since our model can be regarded as a binary classifier, we have also compared our method with other classical binary classifiers. For this paper, we chose three popular representatives: SVM \cite{chang2011libsvm}, logistic regression (LR) \cite{fan2008liblinear}, and Adaboost \cite{freund1999short}. Table \ref{table:bc} demonstrates that the performance of our method GaussianFace-BC is much better than those of the other classifiers. Furthermore, these experimental results demonstrates the effectiveness of the multi-task learning constraint. For example, our GaussianFace-BC has about 7.5\% improvement when all four source-domain datasets are used for training, while the best one of the other three binary classifiers has only around 4\% improvement.

\begin{table}\small
\begin{center}
\begin{tabular}{c|c|c|c|c|c}
  \hline
  The Number of SD& 0 & 1 & 2 & 3 & 4 \\
  \hline\hline
  SVM \cite{chang2011libsvm}& 83.21 & 84.32 & 85.06 & 86.43 & 87.31  \\
  LR \cite{fan2008liblinear}&81.14  & 81.92 & 82.65 & 83.84 & 84.75 \\
  Adaboost \cite{freund1999short}& 82.91 & 83.62 & 84.80 & 86.30 & 87.21 \\
  \textbf{GaussianFace-BC}& \textbf{86.25} & \textbf{88.24} & \textbf{90.01} & \textbf{92.22} & \textbf{93.73} \\
  \hline
\end{tabular}
\end{center}
\caption{The accuracy rate ($\%$) of our methods as a binary classifier and other competing methods on LFW using the increasing number of source-domain datasets.}\label{table:bc}
\end{table}

\begin{table}\small
\begin{center}
\begin{tabular}{c|c|c|c|c|c}
  \hline
  The Number of SD& 0 & 1 & 2 & 3 & 4 \\
  \hline\hline
   K-means \cite{hussain2012face}& 84.71 & 85.20 & 85.74 & 86.81 & 87.68 \\
  RP Tree \cite{dasgupta2009random}& 85.11 & 85.70 & 86.45 & 87.52 & 88.34 \\
  GMM \cite{simonyan2004fisher}& 86.63 & 87.02 & 87.58 & 88.60 & 89.21 \\
  \textbf{GaussianFace-FE} & \textbf{89.33} & \textbf{91.04} & \textbf{93.31} & \textbf{95.62} & \textbf{97.79} \\
  \hline
\end{tabular}
\end{center}
\caption{The accuracy rate ($\%$) of our methods as a feature extractor and other competing methods on LFW using the increasing number of source-domain datasets.}\label{table:fe}
\end{table}

\subsection{Comparisons with Other Feature Extractors} \label{section:fe}

Our model can also be regarded as a feature extractor, which is implemented by clustering to generate a codebook. Therefore, we evaluate our method by comparing it with three popular clustering methods: K-means \cite{hussain2012face}, Random Projection (RP) tree \cite{dasgupta2009random}, and Gaussian Mixture Model (GMM) \cite{simonyan2004fisher}. Since our method can determine the number of clusters automatically, for fair comparison, all the other methods generate the same number of clusters as ours. As shown in Table \ref{table:fe}, our method GaussianFace-FE significantly outperforms all of the compared approaches, which verifies the effectiveness of our method as a feature extractor. The results have also proved that the multi-task learning constraint is effective. Each time one different type of source-domain dataset is added for training,  the performance can be improved significantly. Our GaussianFace-FE model achieves over 8\% improvement when the number of SD varies from 0 to 4, which is much higher than the $\sim$3\% improvement of the other methods.

\subsection{Comparison with the state-of-art Methods}

\begin{figure}[t]
  \centering
  \includegraphics[width=\linewidth]{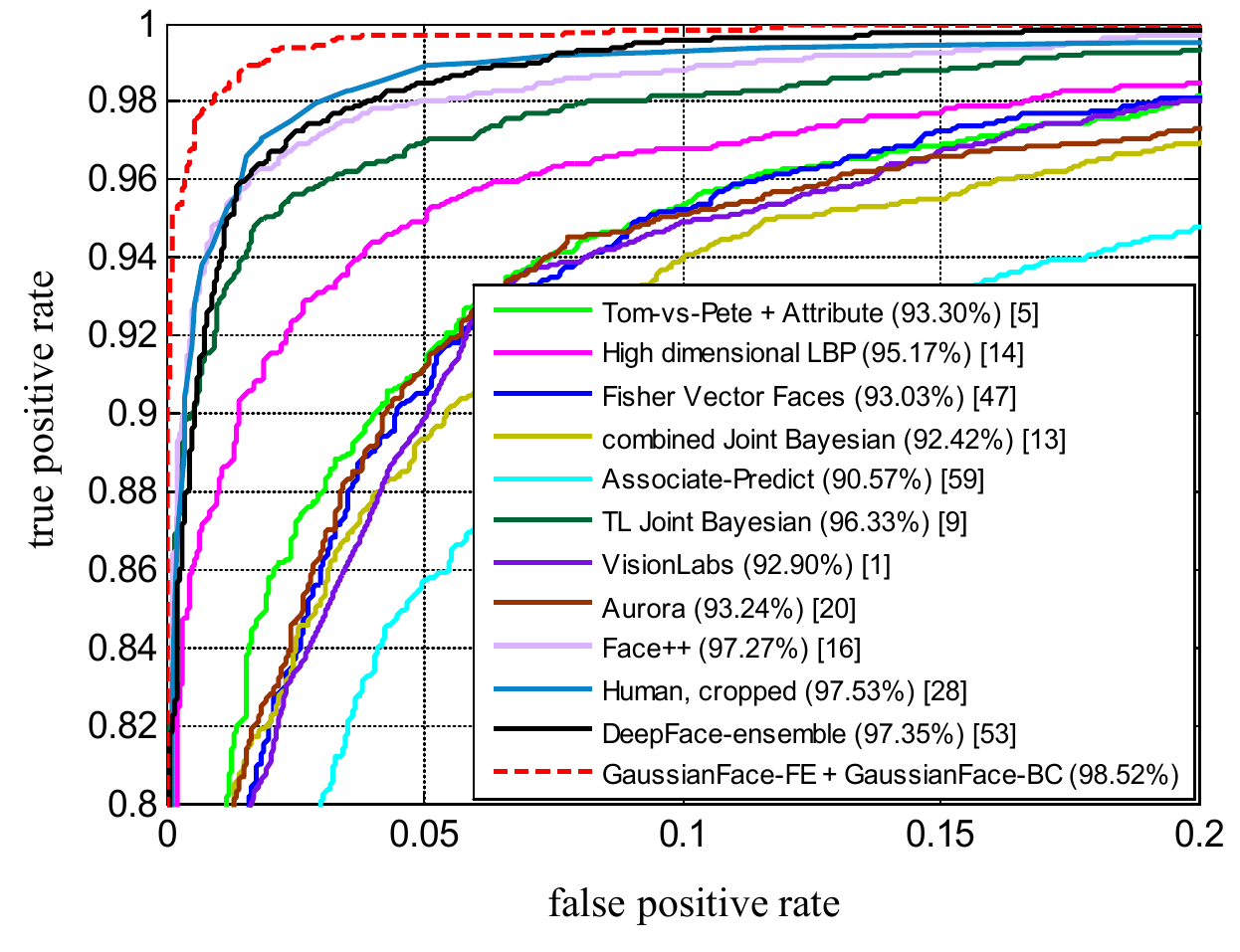}
  \caption{The ROC curve on LFW. Our method achieves the best performance, beating human-level performance.
  }\label{fig:roc}
\end{figure}

\begin{figure}[t]
  \centering
  \includegraphics[width=\linewidth]{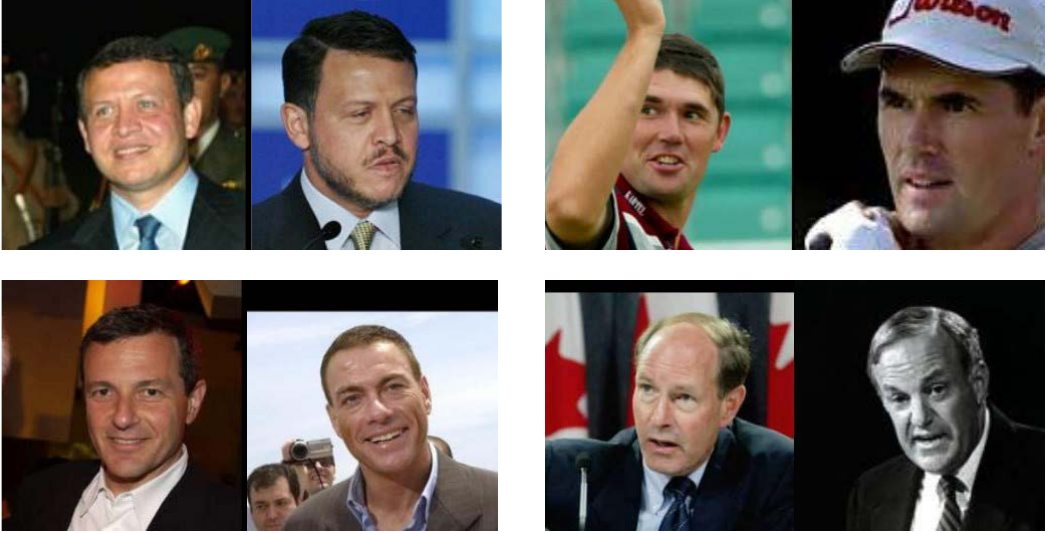}
  \caption{The two rows present examples of matched and mismatched pairs respectively from LFW that were incorrectly classified by the GaussianFace model.
  }\label{fig:matched_mismatched_samples}
\end{figure}

Motivated by the appealing performance of both GaussianFace-BC and GaussianFace-FE, we further combine them for face verification. Specifically, after facial features are extracted using GaussianFace-FE, GaussianFace-BC \footnote[4]{Here, the GaussianFace BC is trained with the extracted high-dimensional features using GaussianFace-FE.} is used to make the final decision. Figure \ref{fig:roc} shows the results of this combination compared with state-of-the-art methods \cite{Cao2013Transfer,berg2012tom,Dong2013Blessing,simonyan2004fisher,chen2012bayesian,deepFace,yin2011associate,visionlab,liaurora,fan2014learning}. The best published result on the LFW benchmark is 97.35\%, which is achieved by \cite{deepFace}. Our GaussianFace model can improve the accuracy to 98.52\%, which for the first time beats the human-level performance (97.53\%, cropped) \cite{kumar2009attribute}. Figure \ref{fig:matched_mismatched_samples} presents some example pairs that were always incorrectly classified by our model. Obviously, even for humans, it is also difficult to verify some of them. Here, we emphasize that the centers of patches, instead of the accurate and dense facial landmarks like \cite{Cao2013Transfer}, are utilized to extract multi-scale features in our method. This makes our method simpler and easier to use.

\subsection{Further Validations: Shuffling the Source-Target}

\begin{figure*}
  \centering
  \includegraphics[width=0.95\linewidth]{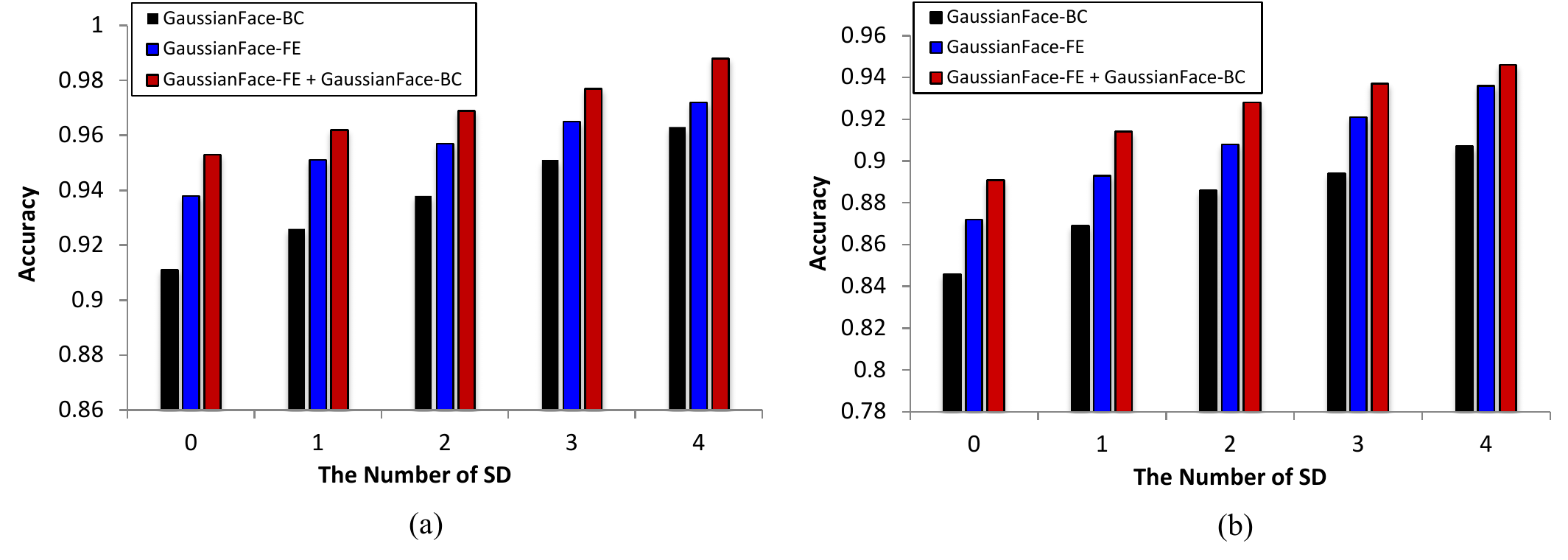}
  \caption{(a) The accuracy rate (\%) of the GaussianFace model on Multi-PIE. (b) The accuracy rate (\%) of the GaussianFace model on MORPH.
  }\label{fig:additional_results}
\end{figure*}

To further prove the validity of our model, we also consider to treat Multi-PIE and MORPH respectively as the target-domain dataset and the others as the source-domain datasets. The target-domain dataset is split into two mutually exclusive parts: one consisting of 20,000 matched pairs and 20,000 mismatched pairs is used for training, the other is used for test. In the test set, similar to the protocol of LFW, we select 10 mutually exclusive subsets, where each subset consists of 300 matched pairs and 300 mismatched pairs. The experimental results are presented in Figure \ref{fig:additional_results}. Each time one dataset is added to the training set, the performance can be improved, even though the types of data are very different in the training set.

\section{General Discussion}

There is an implicit belief among many psychologists and computer scientists that human face verification abilities are currently beyond existing computer-based face verification algorithms \cite{o2006predicting}. This belief, however, is supported more by anecdotal impression than by scientific evidence. By contrast, there have already been a number of papers comparing human and computer-based face verification performance \cite{adler2007comparing,tang2004face,o2007face,phillips2014comparison,o2012comparing,bruce1998comparisons}. It has been shown that the best current face verification algorithms perform better than humans in the good and moderate conditions. So, it is really not that difficult to beat human performance in some specific scenarios.

As pointed out by \cite{o2012comparing,sinha2005face}, humans and computer-based algorithms have different strategies in face verification. Indeed, by contrast to performance with unfamiliar faces, human face verification abilities for familiar faces are relatively robust to changes in viewing parameters such as illumination and pose. For example, Bruce \cite{bruce1982changing} found human recognition memory for unfamiliar faces dropped substantially when there were changes in viewing parameters. Besides, humans can take advantages of non-face configurable information from the combination of the face and body (e.g., neck, shoulders). It has also been examined in \cite{kumar2009attribute}, where the human performance drops from 99.20\% (tested using the original LFW images) to 97.53\% (tested using the cropped LFW images). Hence, the experiments comparing human and computer performance may not show human face verification skill at their best, because humans were asked to match the cropped faces of people previously unfamiliar to them. To the contrary, those experiments can fully show the performance of computer-based face verification algorithms. First, the algorithms can exploit information from enough training images with variations in all viewing parameters to improve face verification performance, which is similar to information humans acquire in developing face verification skills and in becoming familiar with individuals. Second, the algorithms might exploit useful, but subtle, image-based detailed information that give them a slight, but consistent, advantage over humans.

Therefore, surpassing the human-level performance may only be symbolically significant. In reality, a lot of challenges still lay ahead. To compete successfully with humans, more factors such as the robustness to familiar faces and the usage of non-face information, need to be considered in developing future face verification algorithms.

\section{Conclusion and Future Work}

This paper presents a principled Multi-Task Learning approach based on Discriminative Gaussian Process Latent Variable Model, named \emph{GaussianFace}, for face verification by including a computationally more efficient equivalent form of KFDA and the multi-task learning constraint to the DGPLVM model. We use Gaussian Processes approximation and anchor graphs to speed up the inference and prediction of our model. Based on the GaussianFace model, we propose two different approaches for face verification. Extensive experiments on challenging datasets validate the efficacy of our model. The GaussianFace model finally surpassed human-level face verification accuracy, thanks to exploiting additional data from multiple source-domains to improve the generalization performance of face verification in the target-domain and adapting automatically to complex face variations.

Although several techniques such as the Laplace approximation and anchor graph are introduced to speed up the process of inference and prediction in our GaussianFace model, it still takes a long time to train our model for the high performance. In addition, large memory is also necessary. Therefore, for specific application, one needs to balance the three dimensions: memory, running time, and performance. Generally speaking, higher performance requires more memory and more running time. In the future, the issue of running time can be further addressed by the distributed parallel algorithm or the GPU implementation of large matrix inversion. To address the issue of memory, some online algorithms for training need to be developed. Another more intuitive method is to seek a more efficient sparse representation for the large covariance matrix.

\section*{Acknowledgements}

We would like to thank Deli Zhao and Chen Change Loy for their insightful discussions. This work is partially supported by "CUHK Computer Vision Cooperation" grant from Huawei, and by the General Research Fund sponsored by the Research Grants Council of Hong Kong (Project No.CUHK 416510 and 416312) and Guangdong Innovative Research Team Program (No.201001D0104648280).

{
\bibliographystyle{ieee}
\bibliography{GaussianFace}

\begin{thebibliography}{10}\itemsep=-1pt

\bibitem{visionlab}
Visionlabs.
\newblock In {\em Website: http://www.visionlabs.ru/face-recognition}.

\bibitem{adler2007comparing}
A.~Adler and M.~E. Schuckers.
\newblock Comparing human and automatic face recognition performance.
\newblock {\em IEEE Transactions on Systems, Man, and Cybernetics, Part B:
  Cybernetics}, 37(5):1248--1255, 2007.

\bibitem{ahonen2006face}
T.~Ahonen, A.~Hadid, and M.~Pietikainen.
\newblock Face description with local binary patterns: Application to face
  recognition.
\newblock {\em TPAMI}, 28(12):2037--2041, 2006.

\bibitem{ben2002support}
A.~Ben-Hur, D.~Horn, H.~T. Siegelmann, and V.~Vapnik.
\newblock Support vector clustering.
\newblock {\em JMLR}, 2, 2002.

\bibitem{berg2012tom}
T.~Berg and P.~N. Belhumeur.
\newblock Tom-vs-pete classifiers and identity-preserving alignment for face
  verification.
\newblock In {\em BMVC}, volume~1, page~5, 2012.

\bibitem{bonilla2008multi}
E.~Bonilla, K.~M. Chai, and C.~Williams.
\newblock Multi-task gaussian process prediction.
\newblock In {\em NIPS}, 2008.

\bibitem{bruce1982changing}
V.~Bruce.
\newblock Changing faces: Visual and non-visual coding processes in face
  recognition.
\newblock {\em British Journal of Psychology}, 73(1):105--116, 1982.

\bibitem{bruce1998comparisons}
V.~Bruce, P.~J. Hancock, and A.~M. Burton.
\newblock Comparisons between human and computer recognition of faces.
\newblock In {\em Automatic Face and Gesture Recognition}, pages 408--413,
  1998.

\bibitem{Cao2013Transfer}
X.~Cao, D.~Wipf, F.~Wen, and G.~Duan.
\newblock A practical transfer learning algorithm for face verification.
\newblock In {\em ICCV}. 2013.

\bibitem{cao2010face}
Z.~Cao, Q.~Yin, X.~Tang, and J.~Sun.
\newblock Face recognition with learning-based descriptor.
\newblock In {\em CVPR}, pages 2707--2714, 2010.

\bibitem{chai2010multi}
K.~M. Chai.
\newblock Multi-task learning with gaussian processes.
\newblock The University of Edinburgh, 2010.

\bibitem{chang2011libsvm}
C.-C. Chang and C.-J. Lin.
\newblock Libsvm: a library for support vector machines.
\newblock {\em ACM TIST}, 2(3):27, 2011.

\bibitem{chen2012bayesian}
D.~Chen, X.~Cao, L.~Wang, F.~Wen, and J.~Sun.
\newblock Bayesian face revisited: A joint formulation.
\newblock In {\em ECCV}, pages 566--579. 2012.

\bibitem{Dong2013Blessing}
D.~Chen, X.~Cao, F.~Wen, and J.~Sun.
\newblock Blessing of dimensionality: High-dimensional feature and its
  efficient compression for face verification.
\newblock In {\em CVPR}. 2013.

\bibitem{dasgupta2009random}
S.~Dasgupta and Y.~Freund.
\newblock Random projection trees for vector quantization.
\newblock {\em IEEE Transactions on Information Theory}, 55(7):3229--3242,
  2009.

\bibitem{fan2014learning}
H.~Fan, Z.~Cao, Y.~Jiang, Q.~Yin, and C.~Doudou.
\newblock Learning deep face representation.
\newblock {\em arXiv preprint arXiv:1403.2802}, 2014.

\bibitem{fan2008liblinear}
R.-E. Fan, K.-W. Chang, C.-J. Hsieh, X.-R. Wang, and C.-J. Lin.
\newblock Liblinear: A library for large linear classification.
\newblock {\em JMLR}, 9:1871--1874, 2008.

\bibitem{freund1999short}
Y.~Freund, R.~Schapire, and N.~Abe.
\newblock A short introduction to boosting.
\newblock {\em Journal-Japanese Society For Artificial Intelligence},
  14(771-780):1612, 1999.

\bibitem{gross2010multi}
R.~Gross, I.~Matthews, J.~Cohn, T.~Kanade, and S.~Baker.
\newblock Multi-pie.
\newblock {\em Image and Vision Computing}, 28(5):807--813, 2010.

\bibitem{liaurora}
T.~Heseltine, P.~Szeptycki, J.~Gomes, M.~Ruiz, and P.~Li.
\newblock Aurora face recognition technical report: Evaluation of algorithm
  “aurora-c-2014-1” on labeled faces in the wild.

\bibitem{higham1996accuracy}
N.~J. Higham.
\newblock {\em Accuracy and Stability of Numberical Algorithms}.
\newblock Number~48. Siam, 1996.

\bibitem{Huang2012Learning}
G.~Huang, H.~Lee, and E.~Learned-Miller.
\newblock Learning hierarchical representations for face verification with
  convolutional deep belief networks.
\newblock In {\em CVPR}, 2012.

\bibitem{LFWTech}
G.~B. Huang, M.~Ramesh, T.~Berg, and E.~Learned-Miller.
\newblock Labeled faces in the wild: A database for studying face recognition
  in unconstrained environments.
\newblock Technical Report 07-49, University of Massachusetts, Amherst, 2007.

\bibitem{hussain2012face}
S.~U. Hussain, T.~Napol{\'e}on, F.~Jurie, et~al.
\newblock Face recognition using local quantized patterns.
\newblock In {\em BMVC}, 2012.

\bibitem{kim2006appearance}
H.-C. Kim, D.~Kim, Z.~Ghahramani, and S.~Y. Bang.
\newblock Appearance-based gender classification with gaussian processes.
\newblock {\em Pattern Recognition Letters}, 27(6):618--626, 2006.

\bibitem{kim2007clustering}
H.-C. Kim and J.~Lee.
\newblock Clustering based on gaussian processes.
\newblock {\em Neural computation}, 19(11), 2007.

\bibitem{kim2006optimal}
S.-J. Kim, A.~Magnani, and S.~Boyd.
\newblock Optimal kernel selection in kernel fisher discriminant analysis.
\newblock In {\em ICML}, pages 465--472, 2006.

\bibitem{kumar2009attribute}
N.~Kumar, A.~C. Berg, P.~N. Belhumeur, and S.~K. Nayar.
\newblock Attribute and simile classifiers for face verification.
\newblock In {\em ICCV}, pages 365--372, 2009.

\bibitem{lawrence2003gaussian}
N.~D. Lawrence.
\newblock Gaussian process latent variable models for visualisation of high
  dimensional data.
\newblock In {\em NIPS}, volume~2, page~5, 2003.

\bibitem{leen2011focused}
G.~Leen, J.~Peltonen, and S.~Kaski.
\newblock Focused multi-task learning using gaussian processes.
\newblock In {\em Machine Learning and Knowledge Discovery in Databases}, pages
  310--325. 2011.

\bibitem{liprobabilistic}
H.~Li, G.~Hua, Z.~Lin, J.~Brandt, and J.~Yang.
\newblock Probabilistic elastic matching for pose variant face verification.
\newblock In {\em CVPR}. 2013.

\bibitem{li2009nonparametric}
Z.~Li, D.~Lin, and X.~Tang.
\newblock Nonparametric discriminant analysis for face recognition.
\newblock {\em TPAMI}, 31(4):755--761, 2009.

\bibitem{li2005nonparametric}
Z.~Li, W.~Liu, D.~Lin, and X.~Tang.
\newblock Nonparametric subspace analysis for face recognition.
\newblock In {\em CVPR}, volume~2, pages 961--966, 2005.

\bibitem{liu2002gabor}
C.~Liu and H.~Wechsler.
\newblock Gabor feature based classification using the enhanced fisher linear
  discriminant model for face recognition.
\newblock {\em TIP}, 2002.

\bibitem{liu2010large}
W.~Liu, J.~He, and S.-F. Chang.
\newblock Large graph construction for scalable semi-supervised learning.
\newblock In {\em ICML}, pages 679--686, 2010.

\bibitem{lowe2004distinctive}
D.~G. Lowe.
\newblock Distinctive image features from scale-invariant keypoints.
\newblock {\em IJCV}, 60(2):91--110, 2004.

\bibitem{moghaddam2000bayesian}
B.~Moghaddam, T.~Jebara, and A.~Pentland.
\newblock Bayesian face recognition.
\newblock {\em Pattern Recognition}, 33(11):1771--1782, 2000.

\bibitem{o2012comparing}
A.~J. O'Toole, X.~An, J.~Dunlop, V.~Natu, and P.~J. Phillips.
\newblock Comparing face recognition algorithms to humans on challenging tasks.
\newblock {\em ACM Transactions on Applied Perception}, 9(4):16, 2012.

\bibitem{o2006predicting}
A.~J. O’Toole, F.~Jiang, D.~Roark, and H.~Abdi.
\newblock Predicting human performance for face recognition.
\newblock {\em Face Processing: Advanced Methods and Models. Elsevier,
  Amsterdam}, 2006.

\bibitem{o2007face}
A.~J. O'Toole, P.~J. Phillips, F.~Jiang, J.~Ayyad, N.~P{\'e}nard, and H.~Abdi.
\newblock Face recognition algorithms surpass humans matching faces over
  changes in illumination.
\newblock {\em TPAMI}, 29(9):1642--1646, 2007.

\bibitem{phillips2014comparison}
P.~J. Phillips and A.~J. O'Toole.
\newblock Comparison of human and computer performance across face recognition
  experiments.
\newblock {\em Image and Vision Computing}, 32(1):74--85, 2014.

\bibitem{carl2006gp}
C.~E. Rasmussen and C.~K.~I. Williams.
\newblock Gaussian processes for machine learning.
\newblock 2006.

\bibitem{ricanek2006morph}
K.~Ricanek and T.~Tesafaye.
\newblock Morph: A longitudinal image database of normal adult age-progression.
\newblock In {\em Automatic Face and Gesture Recognition}, pages 341--345,
  2006.

\bibitem{rudovic2010coupled}
O.~Rudovic, I.~Patras, and M.~Pantic.
\newblock Coupled gaussian process regression for pose-invariant facial
  expression recognition.
\newblock In {\em ECCV}. 2010.

\bibitem{salakhutdinov2007using}
R.~Salakhutdinov and G.~E. Hinton.
\newblock Using deep belief nets to learn covariance kernels for gaussian
  processes.
\newblock In {\em NIPS}, 2007.

\bibitem{seo2011face}
H.~J. Seo and P.~Milanfar.
\newblock Face verification using the lark representation.
\newblock {\em TIFS}, 6(4):1275--1286, 2011.

\bibitem{simonyan2004fisher}
K.~Simonyan, O.~M. Parkhi, A.~Vedaldi, and A.~Zisserman.
\newblock Fisher vector faces in the wild.
\newblock {\em IJCV}, 60(2):91--110, 2004.

\bibitem{sinha2005face}
P.~Sinha, B.~Balas, Y.~Ostrovsky, and R.~Russell.
\newblock Face recognition by humans: 20 results all computer vision
  researchers should know about.
\newblock {\em Department of Brain and Cognitive Sciences, MIT, Cambridge, MA},
  2005.

\bibitem{skolidis2011bayesian}
G.~Skolidis and G.~Sanguinetti.
\newblock Bayesian multitask classification with gaussian process priors.
\newblock {\em IEEE Transactions on Neural Networks}, 22(12), 2011.

\bibitem{Sun2013Hybrid}
Y.~Sun, X.~Wang, and X.~Tang.
\newblock Hybrid deep learning for face verification.
\newblock In {\em ICCV}. 2013.

\bibitem{sun2014}
Y.~Sun, X.~Wang, and X.~Tang.
\newblock Deep learning face representation from predicting 10,000 classes.
\newblock In {\em CVPR}, 2014.

\bibitem{taigman2009multiple}
Y.~Taigman, L.~Wolf, and T.~Hassner.
\newblock Multiple one-shots for utilizing class label information.
\newblock In {\em BMVC}, pages 1--12, 2009.

\bibitem{deepFace}
Y.~Taigman, M.~Yang, M.~Ranzato, and L.~Wolf.
\newblock {DeepFace: Closing the Gap to Human-Level Performance in Face
  Verification}.
\newblock {\em CVPR, 2014}.

\bibitem{tang2004face}
X.~Tang and X.~Wang.
\newblock Face sketch recognition.
\newblock {\em IEEE Transactions on Circuits and Systems for Video Technology},
  14(1):50--57, 2004.

\bibitem{torralba2011unbiased}
A.~Torralba and A.~A. Efros.
\newblock Unbiased look at dataset bias.
\newblock In {\em CVPR}, pages 1521--1528, 2011.

\bibitem{turk1991face}
M.~A. Turk and A.~P. Pentland.
\newblock Face recognition using eigenfaces.
\newblock In {\em CVPR}, pages 586--591, 1991.

\bibitem{urtasun2007discriminative}
R.~Urtasun and T.~Darrell.
\newblock Discriminative gaussian process latent variable model for
  classification.
\newblock In {\em ICML}, pages 927--934, 2007.

\bibitem{wright2009implicit}
J.~Wright and G.~Hua.
\newblock Implicit elastic matching with random projections for pose-variant
  face recognition.
\newblock In {\em CVPR}, pages 1502--1509, 2009.

\bibitem{yin2011associate}
Q.~Yin, X.~Tang, and J.~Sun.
\newblock An associate-predict model for face recognition.
\newblock In {\em CVPR}, pages 497--504, 2011.

\bibitem{yu2005learning}
K.~Yu, V.~Tresp, and A.~Schwaighofer.
\newblock Learning gaussian processes from multiple tasks.
\newblock In {\em ICML}, pages 1012--1019, 2005.

\bibitem{zhang2010multi}
Y.~Zhang and D.-Y. Yeung.
\newblock Multi-task warped gaussian process for personalized age estimation.
\newblock In {\em CVPR}, pages 2622--2629, 2010.

\bibitem{Zhu2013Deep}
Z.~Zhu, P.~Luo, X.~Wang, and X.~Tang.
\newblock Deep learning identity preserving face space.
\newblock In {\em ICCV}. 2013.

\bibitem{ping2014}
Z.~Zhu, P.~Luo, X.~Wang, and X.~Tang.
\newblock Recover canonical-view faces in the wild with deep neural networks.
\newblock {\em arXiv:1404.3543}, 2014.

\end{thebibliography}
}

\end{document}